\documentclass[letterpaper]{article} 
\usepackage{aaai25}  
\usepackage{times}  
\usepackage{helvet}  
\usepackage{courier}  
\usepackage[hyphens]{url}  
\usepackage{graphicx} 
\usepackage{natbib}  
\usepackage{caption} 
\usepackage{algorithm}
\usepackage{algorithmic}

\usepackage{multirow} 
\usepackage{amsmath}
\usepackage{amssymb}
\usepackage{bm}
\usepackage{pifont}
\usepackage[american]{babel}
\usepackage{microtype}

\usepackage{xcolor}
\usepackage{color}

\usepackage{subcaption}
\definecolor{aaaiblue}{rgb}{0,0,0}
\usepackage{newfloat}
\usepackage{listings}
\usepackage{booktabs}

\DeclareCaptionStyle{ruled}{labelfont=normalfont,labelsep=colon,strut=off} 
\floatstyle{ruled}
\newfloat{listing}{tb}{lst}{}
\floatname{listing}{Listing}
\newcommand{\argmin}{\mathop{\rm arg~min}\limits}

\newcommand{\etal}{\textrm{et~al.\,}}

\title{\textsc{BrainGuard}: Privacy-Preserving Multisubject \\ Image Reconstructions from Brain Activities}
\author{Zhibo Tian\textsuperscript{\rm 1}, 
Ruijie Quan\textsuperscript{\rm 2}, 
Fan Ma\textsuperscript{\rm 3},
Kun Zhan\textsuperscript{\rm 1}\thanks{Corresponding Author.}, 
Yi Yang\textsuperscript{\rm 3}}
\affiliations{
\textsuperscript{\rm 1}School of Information Science and Engineering, Lanzhou University\\
\textsuperscript{\rm 2}College of Computing and Data Science, Nanyang Technological University\\
\textsuperscript{\rm 3}College of Computer Science and Technology, Zhejiang University\\
https://github.com/kunzhan/BrainGuard
}
\begin{document}\maketitle
\begin{abstract}
Reconstructing perceived images from human brain activity forms a crucial link between human and machine learning through Brain-Computer Interfaces. Early methods primarily focused on training separate models for each individual to account for individual variability in brain activity, overlooking valuable cross-subject commonalities. Recent advancements have explored multisubject methods, but these approaches face significant challenges, particularly in data privacy and effectively managing individual variability. To overcome these challenges, we introduce \textsc{BrainGuard}, a privacy-preserving collaborative training framework designed to enhance image reconstruction from multisubject fMRI data while safeguarding individual privacy. \textsc{BrainGuard} employs a collaborative global-local architecture where individual models are trained on each subject's local data and operate in conjunction with a shared global model that captures and leverages cross-subject patterns. This architecture eliminates the need to aggregate fMRI data across subjects, thereby ensuring privacy preservation. To tackle the complexity of fMRI data, \textsc{BrainGuard} integrates a hybrid synchronization strategy, enabling individual models to dynamically incorporate parameters from the global model. By establishing a secure and collaborative training environment, \textsc{BrainGuard} not only protects sensitive brain data but also improves the image reconstructions accuracy. Extensive experiments demonstrate that \textsc{BrainGuard} sets a new benchmark in both high-level and low-level metrics, advancing the state-of-the-art in brain decoding through its innovative design.
\end{abstract}
\section{Introduction}
Deciphering brain activities and recovering encoded information represents a fundamental objective in the field of cognitive neuroscience~\cite{chen2023seeing,bullmore2009complex,parthasarathy2017neural,beliy2019voxels}. Functional Magnetic Resonance Imaging
(fMRI) data~\cite{kwong1992dynamic} is commonly used to recover visual information and plays a crucial role in delineating brain activity by observing fluctuations in blood oxygenation level. Initially, methods for interpreting brain activity from fMRI mainly involved image classification~\cite{kamitani2005decoding,cox2003functional}.
With the integration of deep generative models, the focus has shifted towards more sophisticated fMRI-to-Image (f2I) reconstruction approaches~\cite{shen2019deep,beliy2019voxels}. However, the inherent variability in fMRI across individuals presents significant challenges, as individuals exhibit unique brain activation patterns in response to the same visual stimulus~\cite{chen2023seeing}.
\begin{figure}
\centering
\includegraphics[width=\linewidth]{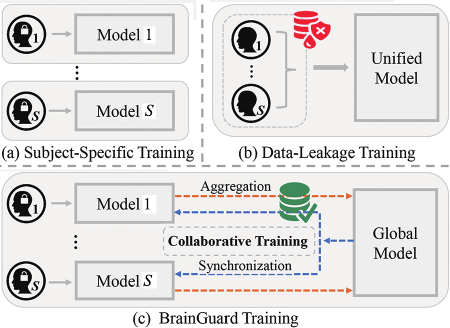}
\caption{
(a) Early subject-specific methods require separate training for each individual using their respective fMRI, overlooking intersubject commonalities.
(b) Recent multisubject methods that combine all subjects' fMRI for training pose substantial privacy concerns.
(c) \textsc{BrainGuard} captures intersubject commonalities while preserving data privacy.
}
\label{fig:motivation}
\vspace{-10pt}
\end{figure}

Earlier f2I reconstruction methods employed a subject-specific approach, where distinct models are developed for each individual’s fMRI to address the significant variance in brain activity across different subjects (Fig.~\ref{fig:motivation}(a))~\cite{lin2022mind,scotti2023reconstructing,gu2022decoding,chen2023seeing,takagi2023high}. However, this approach is inherently limited by its focus on individual differences, which results in the neglect of valuable insights that could be obtained from commonalities across multisubject fMRI. Recent advancements increasingly focus on multisubject or cross-subject methods (Fig.~\ref{fig:motivation}(b))~\cite{quan2024psychometry,scotti2024mindeye2,wang2024mindbridge}, which aims to train a unified model capable of reconstructing images from the fMRI of multiple individuals. These methods incorporate simple subject-specific layers, such as ridge regression, to project fMRI of different subjects into a common embedding space, thus accounting for individual differences while training on combined dataset.

\begin{figure}
\centering
\includegraphics[width=\linewidth]{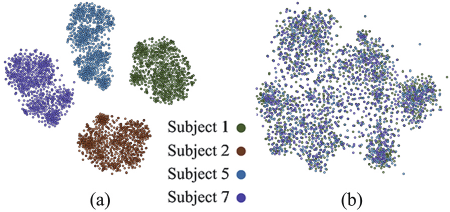}
\vspace{-15pt}
\caption{(a) t-SNE visualization of the embeddings output from the subject-specific layers in MindBridge~\cite{wang2024mindbridge}. (b) represents ones from \textsc{BrainGuard} models. Both are visualized upon the NSD~\cite{allen2022massive} \texttt{test} set.}\label{fig:tsne}
\vspace{-15pt}
\end{figure}

However, two primary challenges remain. \textbf{First}, exiting multisubject methods that combine all subjects' fMRI for training poses substantial privacy concerns. Brain data is highly personal and confidential, and aggregating it without robust privacy protections is problematic, especially considering that the data often originates from multiple organizations. \textbf{Second}, the subject-specific layers employed in these methods are inadequate for fully addressing the complex differences in fMRI among individuals. As illustrated in Fig.~\ref{fig:tsne}(a), the embedding output from the subject-specific layers in MindBridge~\cite{wang2024mindbridge} fails to effectively map the fMRI to a common embedding space. These simplistic mappings struggle to capture the distinctive neural patterns of each subject, thus limiting the model’s ability to generalize effectively across diverse data.

In response to these challenges, we introduce \textsc{BrainGuard}, the first privacy-preserving collaborative training framework designed for multisubject f2I reconstruction (Fig.~\ref{fig:motivation}(c)). \textsc{BrainGuard} adopts a global-local model, consisting of individual models for each subject and a global model that captures and leverages shared patterns across subjects. During the training phase, the global model integrates the updated parameters from each individual model, which are independently trained on their respective subject's fMRI, and then the individual models dynamically synchronize the merged parameters from the global. This design promotes a distributed and collaborative learning mechanism that balances personalization with generalization, thereby enhancing reconstruction quality while maintaining data privacy.

Furthermore, unlike typical image and text, fMRI reflects complex neural activity and is typically preprocessed into a 1D vector of high-dimensional voxels. This complexity renders standard data processing methods less effective. To address this, we introduce a \textbf{hybrid synchronization strategy} in \textsc{BrainGuard}, specifically designed to meet the unique challenges of fMRI. This strategy synchronizes different parts of the model layers in a hybrid manner, effectively capturing both the distinct neural activity of individual subjects and the shared patterns across subjects by leveraging the specialized capabilities of different model layers. Specifically, it consists of three parts: \ding{182} retention for foundational layers, \ding{183} global alignment for intermediate layers, and \ding{184} adaptive tuning for advanced layers. The foundational layers retain the subject-specific features inherent in fMRI, preserving the individual variability that is crucial for accurate modeling. Then, the intermediate layers are synchronized with the global model to integrate shared insights, which helps generalize across different subjects. Finally, the advanced layers are adaptively tuned using a Dynamic Fusion Learner (DFL) module, which selectively incorporates global model parameters into the individual models. This hybrid strategy is specifically designed to balance individual uniqueness and commonalities, thereby enhancing the model's generalization ability across diverse subjects while safeguarding the privacy of sensitive data.

\textsc{BrainGuard} enjoys several attractive qualities:
\textbf{First,} \textsc{BrainGuard} adeptly balances individual uniqueness and commonalities across subjects, enhancing the model's generalization capabilities (Fig.~\ref{fig:tsne}(b)) while preserving the specificity of individual subject data.
\textbf{Second,} the hybrid synchronization strategy within \textsc{BrainGuard} ensures that individual models dynamically incorporate broader dataset patterns, thereby improving the precision and personalization of image reconstructions.
\textbf{Third,} \textsc{BrainGuard} ensures the privacy and security of sensitive fMRI by facilitating a collaborative learning framework, allowing for collaborative training without the need for direct data sharing, thus upholding stringent data confidentiality standards.

In summary, our contributions are threefold:
\begin{itemize}
    \item To the best of our knowledge, this is the first method to propose a privacy-preserving approach for reconstructing images from fMRI. We introduce \textsc{BrainGuard}, a framework that captures both intersubject commonalities and individual variabilities while preserving data privacy.
    \item \textsc{BrainGuard} incorporates a hybrid synchronization strategy, enabling each individual model to align with its subject's unique objectives and dynamically assimilate and apply relevant patterns identified by the global model.
    \item Experiments show \textsc{BrainGuard}’s efficacy and adaptability, highlighting its potential to significantly advance the field of brain decoding while perserving privacy.
\end{itemize}
\section{Related Work}
\subsection{Image Reconstruction from fMRI}
Earlier f2I reconstruction studies relied on handcrafted features to translate fMRI into images~\cite{kay2008identifying,miyawaki2008visual,nishimoto2011reconstructing}. Subsequent advances utilized paired fMRI images and sparse linear regression, significantly improving the association between brain activity and visual imagery. In recent years, researchers made notable progress in reconstructing images from fMRI by mapping brain signals to the latent space of generative adversarial networks (GANs)~\cite{lin2022mind,karras2020analyzing}. The introduction of multimodal vision-language models~\cite{radford2021learning,lu2019vilbert,ramesh2021zero,li2023efficient,zhao2023clip4str,yang2024doraemongpt}, diffusion models~\cite{ho2020denoising,sohl2015deep,song2019generative,song2020score,rombach2022high}, and large-scale fMRI datasets~\cite{van2013wu,horikawa2017generic,chang2019bold5000,allen2022massive} elevated the quality of image reconstruction from fMRI to unprecedented levels~\cite{mai2023brain}. These diffusion model-based methods~\cite{ozcelik2023brain,scotti2023reconstructing} map fMRI signals to CLIP text and image embeddings using subject-specific ridge regression or MLPs, followed by a diffusion model that integrates multiple inputs. Recent advances shifted towards the cross-subject method~\cite{quan2024psychometry, scotti2024mindeye2, wang2024mindbridge, gong2024mindtuner}, which aims to generalize models across various subjects.

Although effective, current multisubject methods primarily utilize simple subject-specific layers, such as ridge regression, which are inadequate for capturing individual variability. Moreover, these methods aggregate data from all subjects, raising significant privacy concerns. Consequently, there is a critical need for models that achieve a balance between personalization and shared representations while ensuring privacy through decentralized approaches. Addressing these challenges serves as the motivation for this study.

\subsection{Federated Learning}
Federated learning (FL) addresses the challenges of statistical heterogeneity in distributed systems. Traditional methods like FedAvg~\cite{mcmahan2017communication}, aim to build a single global model for all clients. However, these methods often fail to meet the specific needs of individual clients due to the diverse nature of their data. Personalized Federated Learning (pFL) shifts the focus to creating models tailored to each client, thereby enhancing performance and relevance~\cite{zhang2023fedala}. Recent advancements in pFL include regularizing local loss functions to prevent overfitting~\cite{li2020federated,yao2020continual,li2021model}, fine-tuning and transfer learning~\cite{fallah2020personalized,chen2020fedhealth,yang2020fedsteg}, and applying meta-learning techniques~\cite{li2020federated,yao2020continual,li2021model}. Knowledge distillation has also been used to transfer knowledge from a global model to personalized models, improving accuracy and efficiency~\cite{li2019fedmd,zhu2021data,lin2020ensemble,he2020group}. In addition, clustering groups clients with similar data distributions, facilitating more effective personalization and reducing computational overhead~\cite{sattler2020clustered,briggs2020federated,ghosh2020efficient}. These methodologies ensure that individual client needs are met while maintaining the fundamental principles of data privacy and security in FL.

This work represents the initial exploration of privacy and security protection for sensitive fMRI data in the context of brain activities image reconstruction. Unlike conventional image or text data, fMRI captures intricate neural activity and is typically preprocessed into high-dimensional representations. To address these challenges, we propose a hybrid synchronization strategy tailored to the unique characteristics of fMRI data, improving the model's capacity to learn both individual-specific and common features.
\section{Methodology}
\subsection*{Task Setup and Notations}
Our goal is to improve image reconstruction by leveraging commonalities across multisubject fMRI while maintaining subject privacy.
The original acquired fMRI is four-dimensional (3D space $\!$+$\!$ time). Typically, the data is averaged across the time dimension, resulting in a three-dimensional representation, which is then flattened into a one-dimensional vector of voxels corresponding to the visual stimuli presented to a healthy subject.
Let $\bm{X}_{s,n}\!\in\!{\mathbb{R}^d}$ represent the extracted fMRI when a RGB image (visual stimuli) is presented to the subject $s\!\in\!\{1,\ldots,S\}$, where $d$ is the number of voxels, and $n\!\in\!{\{1,\ldots,N\}}$ denotes the number of images. During the training phase, $\bm{X}_{s,n\!}$ is transformed into a well-aligned CLIP~\cite{radford2021learning} embedding space through the integration of visual and textual supervision. The CLIP embeddings of the ground-truth image and its corresponding caption text are denoted by $\bm{I}_{s,n}\!\in\!{\mathbb{R}^{v\times{c}}}$ and $\bm{T}_{s,n}\!\in\!{\mathbb{R}^{t\times{c}}}$, respectively.
Here, $v$ and $t$ represent the number of tokens in the CLIP image and text embeddings, while $c$ indicates their dimensions.
During inference, the encoded fMRI embeddings, $\tilde{\bm{I}}_{s,n}$ and $\tilde{\bm{T}}_{s,n}$, serve as conditional inputs to a versatile diffusion model~\cite{xu2023versatile}, which facilitates image reconstruction.

\begin{figure*}[!t]
\centering
\includegraphics[width=0.88\textwidth]{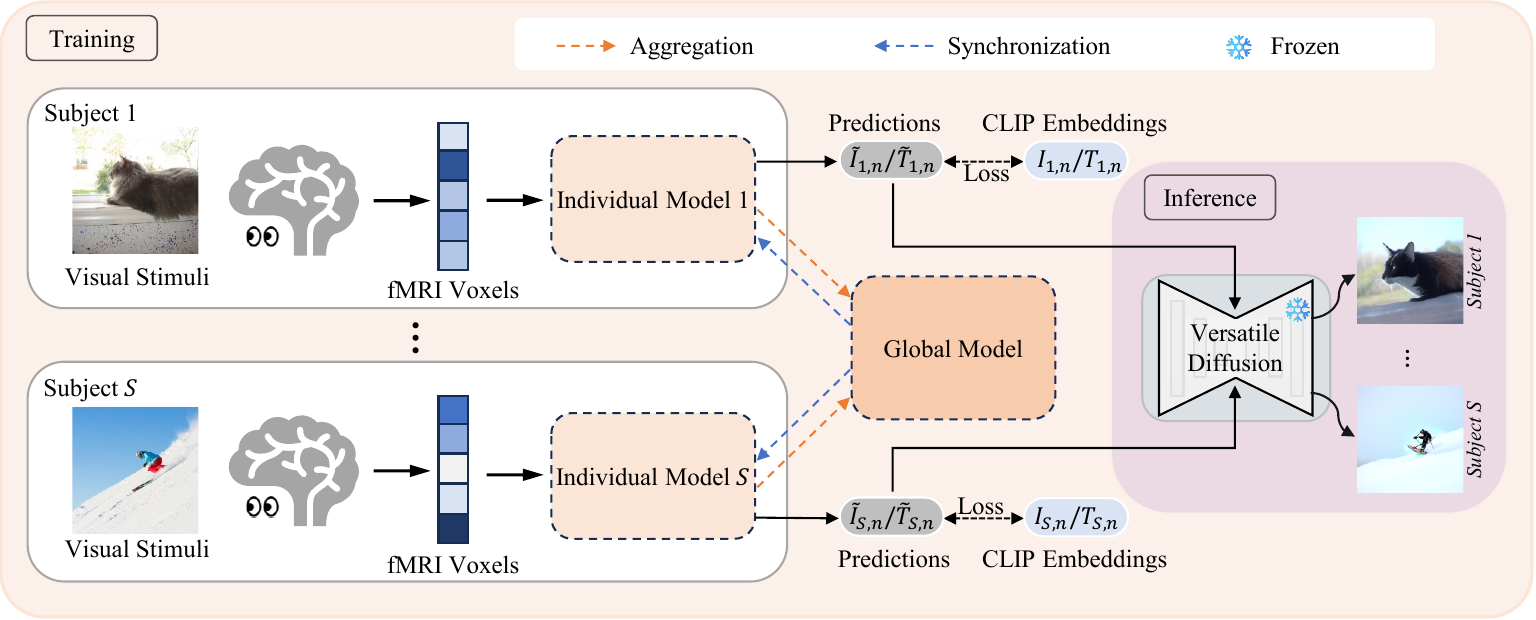}
\caption{
An overview of the \textsc{BrainGuard} training and inference framework (\S\ref{sec_framework}).}\label{fig:framework}
\vspace{-15pt}
\end{figure*}
\subsection*{Method Overview}
\textsc{BrainGuard} requires only a single training session using multisubject data by introducing a collaborative training framework, consisting of individual models and a global model (\S\ref{sec_framework}).
These models are trained in a collaborative approach, where individual and global models engage in bidirectional parameter fusion during their training process. Specifically, the individual models are optimized not only through their respective subject's fMRI training objectives but also via integration with the global model. Conversely, the update of the global model's parameters is informed by the amalgamation of multiple individual models.
To achieve the dynamic parameter fusion process, we devise a hybrid synchronization strategy (\S\ref{subsec_hybrid}). An overview of our framework is illustrated in Fig.~\ref{fig:framework}, and the detailed network architecture and learning objectives is presented in \S\ref{sec:network_arch}.
\subsection{Privacy-Preserving Collaborative Training}\label{sec_framework}
\textsc{BrainGuard} implements a privacy-preserving collaborative training approach designed to protect the privacy of subjects' fMRI while simultaneously leveraging commonalities across different subjects. The training process is cyclical and consists of three essential steps:
\textbf{1) Individual Training:}
At the beginning of each training cycle, \textsc{BrainGuard} trains a distinct model for each subject using their specific fMRI. This approach ensures that sensitive neural data remains localized, thereby eliminating the need for centralization. The individual model, denoted as $\mathtt{f}·s$, is optimized to predict the CLIP image embedding ${\tilde{\bm{I}}}_{s,n}$ and the text embedding ${\tilde{\bm{T}}}_{s,n}$ based on the subject's data.
\textbf{2) Parameters Aggregating:}
 After each training round, the individual models transmit their updated parameters to a global model for aggregation, which aggregates these parameters to capture shared patterns across subjects. Before this aggregation, individual models update their parameters using the Exponential Moving Average (EMA): $\theta'_s = \alpha \theta'_s + (1-\alpha) \theta_s$, where $\alpha$ is the EMA factor, typically set to 0.999. This step aims to smooth parameter changes and enhance individual model stability, ensuring that the subsequent aggregation reflects both refined parameters and shared information. The global model’s parameters $\theta_g$ is updated using $\theta_{\rm g} = \sum_sk_s\theta'_s$, where $k_s$ is a weight coefficient representing each subject's data proportion in the overall dataset, calculated as $k_s=\left| \bm{X}_{s,n} \right|/\sum^{S}_{i=1}\left| \bm{X}_{i,n} \right|$ where $\left| \bm{X}_{s,n} \right|$ denotes the total number of fMRI for subject $s$. This aggregation process enables the global model to effectively capture and integrate shared patterns across subjects.
\textbf{3) Model Synchronizing: }
After the global model is updated, individual models are refined by incorporating the aggregated knowledge from the global through a hybrid synchronization strategy (\S\ref{subsec_hybrid}). This process ensures that each individual benefits from the shared knowledge while preserving its unique adaptations to the specific subject data, thereby achieving a balance between generalization and personalization.
\subsection{Hybrid Synchronization Strategy}\label{subsec_hybrid}
Unlike typical image and text data, fMRI reflects complex neural activity and is typically preprocessed into a one-dimensional vector comprising high-dimensional voxels. This inherent complexity makes standard data processing methods less effective. To address this challenge, inspired by FedALA~\cite{zhang2023fedala}, we employ a hybrid synchronization strategy that balances individual uniqueness with shared insights through the integration of individual and global models. It is implemented through three distinct approaches, each targeting specific layers within the model to optimize different aspects of the learning process based on the unique characteristics of fMRI.

\textbf{Retention for foundational layers.} The fMRI is characterized by highly individualized brain activity patterns, in the lower layers that capture foundational neural signals. These patterns are unique to each subject and vary significantly depending on experimental conditions. To preserve these individualized features, the lower layers of the local models retain their original parameters without updates from the global model. This approach ensures that each local model remains finely tuned to its specific data distribution, improving the accuracy of personalized interpretations of brain activity.

\textbf{Global alignment for intermediate layers.} Despite the individualized nature of foundational layers, intermediate layers of fMRI processing often reveal commonalities across different subjects and experimental conditions, such as shared patterns in brain connectivity. These layers benefit from global insights, which help in aligning the models across different datasets. In this strategy, the intermediate layers of the local models are updated by directly adopting parameters from the corresponding layers in the global model. This promotes a unified feature extraction process that captures the shared characteristics of fMRI, enhancing the model's ability to generalize across diverse subjects and conditions.

\textbf{Adaptive tuning for advanced layers.} The advanced layers of the model are tasked with interpreting complex, high-level features related to cognitive processes and brain network interactions. These layers must balance the nuanced variations in fMRI with broader patterns. To achieve this, we introduce a Dynamic Fusion Learner (DFL) module~\cite{zhang2023fedala}, which adaptively tunes the advanced layers' parameters. This tuning is crucial for integrating both global patterns and local variations, ensuring the model can accurately reflect the cognitive and neural processes underlying the fMRI, thereby improving overall performance and generalization.
Together, these strategies form a cohesive hybrid synchronization approach that effectively balances the unique and shared aspects of individual fMRI, optimizing both individual model performance and global model robustness.

\textbf{Dynamic Fusion Learner Module.} Since the advanced layers of a network generally handle the most abstract high-level features, including complex patterns that are highly task-specific~\cite{yosinski2014transferable, lecun2015deep}, they are critical for fine-tuning in the context of \textsc{BrainGuard}. The distinct characteristics of fMRI further emphasize the importance of these advanced layers for accurately interpreting and responding to individual brain activity. In \textsc{BrainGuard}, we employ a Dynamic Fusion Learner (DFL) module, which performs layer-wise dynamic aggregation of the global and individual model parameters in the advanced layers rather than completely overwriting them. DFL is strategically applied to the advanced $p$ layers, fine-tuning them to reflect individual specificities. This approach seeks to balance the general patterns with the personalized aspects of the individual model through targeted adjustments.

Formally, the parameters of the individual model, $\theta_s$, at iteration $t$, are updated through DFL as follows:
\begin{equation}\small\label{eq:overview of DFL}
\theta_s^t:=\theta_s^{t-1} + (\theta_{\rm g}^{t-1} - \theta_s^{t-1}) \odot{W_s},
\end{equation}
where $\odot$ denotes the Hadamard product. $(\theta_{\rm g}^{t-1} - \theta_s^{t-1})$ accounts for the difference between the global and the individual, while $W_s$ represents the aggregation weights between the global and the individual for specific layer parameters. We apply  element-wise weight clipping~\cite{courbariaux2016binarized,zhang2023fedala}, defined as \(\tau(w) = \max(0, \min(1, w))\), ensuring that \(w \in [0, 1]\) for all \(w \in W_s\). When $\forall w_s \in W_s$, $w_s \equiv 1$, it indicates a scenario where the global completely overwrites the individual models' parameters.

Since we focus exclusively on the advanced \(m\) layers in the model, Eq.~\eqref{eq:overview of DFL} is given by:
\begin{equation}\small\label{eq:update_local}
\theta_s^t:=\theta_s^{t-1} + (\theta_{\rm g}^{t-1} - \theta_s^{t-1}) \odot[\mathbf{1}^{\left| \theta_s \right|-m};W_s^m],
\end{equation}
where $\left|\theta_s\right|$ denotes the total number of layers in the model, and $\mathbf{1}^{\left| \theta_s \right|-m}$ is an array of ones corresponding to the foundational and intermediate layers. $W_s^m$ represents the aggregation weights for the advanced $m$ layers, allowing the model to adapt to subject-specific features of higher level. Initially, each element in $W_s^m$ is set to one and is updated iteratively. The $W_s^m$ is optimized through a gradient-based learning method:
\begin{equation}\small\label{eq:update_weight}
    W_{s}^{m} \leftarrow W_{s}^{m} - \eta \frac{\partial \mathcal{L}(\theta_s; \theta_{\rm g})}{\partial W_{s}^{m}},
\end{equation}
where $\eta$ is the learning rate for weight learning.

By employing DFL, \textsc{BrainGuard} adeptly balances the exploration of intersubject commonalities with the delineation of individual specificities. This balance is crucial for neuroscientific studies and clinical applications where understanding shared brain functions is as important as recognizing subject-specific patterns. This methodology highlights the framework's potential to foster a new paradigm for collaborative yet privacy-preserving neuroimaging studies, wherein collective insights are obtained without compromising individual data confidentiality. 
\subsection{Network Architecture and Training Objectives}\label{sec:network_arch}
\textbf{Network Architecture.}
The network architecture for both the individual and global models is uniform, comprising an initial linear layer, several residual blocks, and two final linear layers (corresponding to the three different synchronization parts). The model projects the fMRI voxels to a shared CLIP latent space, enabling semantic interpretation. During inference, the predicted fMRI embeddings are input into a SOTA diffusion model~\cite{xu2023versatile}, which guides the image reconstruction process. Both the individual and global models can be used to produce the resultant fMRI embeddings.

\textbf{Training Objectives.}
In practice, \textsc{BrainGuard} is optimized by incorporating fMRI as an additional modality, aiming to align the fMRI-derived embeddings more closely with the CLIP space.
Specifically, the individual models employ two types of loss functions to learn CLIP image and text embeddings. The first loss function is the mean squared error, which ensures the accurate prediction of CLIP embeddings:
\begin{equation}\small\label{eq:loss_mse}
\mathcal{L}_{\text{MSE}}(\bm P,\bm Y) = \frac1B \sum_{i=1}^{B} (\bm{p}_i - \bm{y}_i)^2,
\end{equation}
where $\bm{p}$ is the predicted CLIP embedding by the individual model, and $\bm{y}$ is the target embedding in a batch size of $B$. The second loss function, SoftCLIP loss~\cite{scotti2023reconstructing}, facilitates the alignment of fMRI and CLIP embeddings by leveraging contrastive learning to increase positive pair similarity and decrease negative pair similarity. Positive pairs are identified using soft labels derived from the dot product of embeddings within a batch.
\begin{equation}\small\label{eq:loss_softclip}
\begin{aligned}
\mathcal{L}_{\text{SoftCLIP}}(\bm P,\bm Y)
=&-\sum_{i=1}^{B} \sum_{j=1}^{B}\frac{\exp\left( \frac{\bm y_i \cdot \bm y_j}{\tau} \right)}{\sum_{k=1}^{B} \exp\left( \frac{\bm y_i \cdot \bm y_k}{\tau} \right)} \\
&\cdot
\log\left( \frac{\exp\left( \frac{\bm p_i \cdot \bm y_j}{\tau} \right)}{\sum_{k=1}^{B} \exp\left( \frac{\bm p_i \cdot \bm y_k}{\tau} \right)} \right),
\end{aligned}
\end{equation}
where $\tau$ denotes the temperature hyperparameter.

The complete set of losses $\mathcal{L}_s^V$ and $\mathcal{L}_s^T$ for predicting image and text CLIP embeddings includes:
\begin{align}\small
\mathcal{L}_s^V
&= \mathcal{L}_{\text{MSE}}(\tilde{\bm{I}}_{s}, \bm{I}_{s}) + \mathcal{L}_{\text{SoftCLIP}}(\tilde{\bm{I}}_{s}, \bm{I}_{s\!}),\label{eq:loss_total_image}\\
\mathcal{L}_s^T
&= \mathcal{L}_{\text{MSE}}(\tilde{\bm{T}}_{s}, \bm{T}_{s}) + \mathcal{L}_{\text{SoftCLIP}}(\tilde{\bm{T}}_{s}, \bm{T}_{s}).\label{eq:loss_total_text}
\end{align}

The main objective of the entire framework is to minimize the global objective function, balancing the performance of individual models with the integration of commonalities. This is achieved by optimizing both image and text prediction models across all subjects as follows:
\begin{equation}
\bigl\{(\theta_s^\star)\bigr\}_{s=1}^S
=\argmin_{\theta_s}
\sum_{s=1}^S
(\mathcal L_s^V+\mathcal L_s^T)
\,.\label{eq:total_optim_object}
\end{equation}

\begin{table*}[t]
\centering
\resizebox{0.9\textwidth}{!}{
\begin{tabular}{cr|cccc|cccc}
\toprule
~ & & \multicolumn{4}{c}{\texttt{Low-Level}} & \multicolumn{4}{c}{\texttt{High-Level}}\\
\cmidrule{3-10}
\multicolumn{2}{c|}{\multirow{-2}{*}{Methods}}
&{\texttt{PixCorr}}\!~$\uparrow$ &\texttt{SSIM}\!~$\uparrow$ &\texttt{Alex}\,(2)\!~$\uparrow$ &\texttt{Alex}\,(5)\!~$\uparrow$  &{\texttt{Incept}.}\!~$\uparrow$ &\texttt{CLIP}\!~$\uparrow$ &\texttt{EffNet-B}\!~$\downarrow$ &\texttt{SwAV}\!~$\downarrow$  \\
\midrule
\multicolumn{1}{l}
{~Mind-Reader$_{\!}$~\tiny{NeurIPS~2022}}& & $-$ & $-$ & $-$ & $-$  & $78.2\%$ & $-$ & $-$ & $-$ \\
\multicolumn{1}{l}
{~Mind-Vis$_{\!}$~\tiny{CVPR~2023}}& & $.080$ & $.220$ &  $72.1\%$ & $83.2\%$  & $78.8\%$ & $76.2\%$ & $.854$ & $.491$ \\
\multicolumn{1}{l}
{~Takagi~\etal$_{\!}$~\tiny{CVPR~2023}}& & $-$ & $-$ &  $83.0\%$ & $83.0\%$  & $76.0\%$ & $77.0\%$ & $-$ & $-$ \\
\multicolumn{1}{l}
{~Gu~\etal$_{\!}$~\tiny{MIDL~2023}}& & $.150$& ${.325}$ & $-$ & $-$  & $-$ & $-$ & $.862$ & $.465$  \\
\multicolumn{1}{l}
{~Brain-Diffuser$_{\!}$~\tiny{arXiv~2023}}& &$.254$ & $\textbf{.356}$ & ${94.2}\%$ & $ {96.2}\%$ & $87.2\%$ & $91.5\%$ & $.775$ & $.423$ \\
\multicolumn{1}{l}
{~MindEye$_{\!}$~\tiny{NeurIPS~2023}}& &$\underline{.309}$ & $.323$ & $\textbf{94.7}\%$ & $\textbf{97.8}\%$ & ${93.8}\%$ & ${94.1}\%$ & ${.645}$ & ${.367}$ \\
\multicolumn{1}{l}
{~MindBridge$_{\!}$~\tiny{CVPR~2024}}& &$.148$ & $.259$ & ${86.9}\%$ & ${ 95.3}\%$ & ${92.2}\%$ & ${94.3}\%$ & ${.713}$ & ${.413}$ \\
\multicolumn{1}{l}
{~Psychometry$_{\!}$~\tiny{CVPR~2024}}& &$.295$ & $.328$ & $\underline{94.5}\%$ & $\underline{96.8}\%$ & $\underline{94.9}\%$ & $\underline{95.3}\%$ & $\underline{.632}$ & $\underline{.361}$ \\
\cmidrule{1-10}
{\textbf{\textsc{~BrainGuard}}~\footnotesize{(\textbf{\texttt{ours}})}}& & $\textbf{.313}$  & $\underline{.330}$ & $\textbf{94.7}\%$ & $\textbf{97.8}\%$ & $\textbf{96.1}\%$ & $\textbf{96.4}$\% & $\textbf{.624}$ & $\textbf{.353}$  \\
\bottomrule
\end{tabular}}
\caption{\textbf{Quantitative comparison results on NSD \texttt{test} dataset between \textsc{BrainGuard} and previous SOTA methods}. The best result is highlighted in bold, and the second-best result is underlined. Note that all methods employ a \textit{per-subject-per-model} approach for the final inference process, and all metrics are calculated as the average across four subjects.}
 \label{table:all_result}
 \vspace{-10pt}
\end{table*}
\begin{figure*}[!t]
\centering
\includegraphics[width=0.98\textwidth]{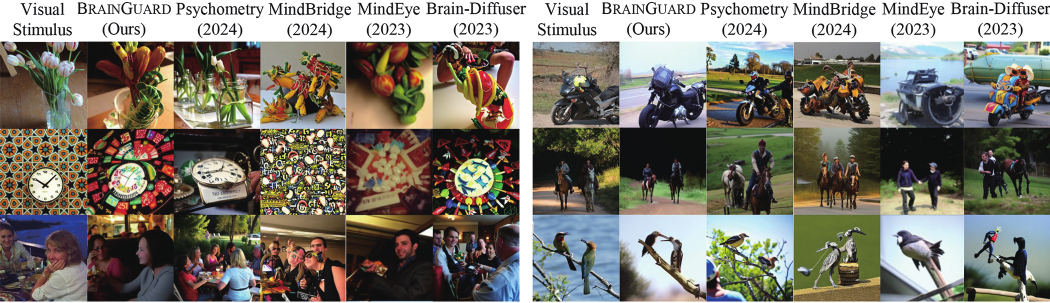}
\caption{\textbf{Qualitative comparisons on the NSD \texttt{test} dataset.} \textsc{BrainGuard} performs a single training session on multisubject fMRI data, demonstrates superior reconstruction accuracy compared to four recent state-of-the-art methods~\cite{quan2024psychometry,wang2024mindbridge,scotti2023reconstructing,ozcelik2023brain}, while effectively preserving data privacy.}\label{fig:all_sub}
\vspace{-15pt}
\end{figure*}

\section{Experiments}
\subsection*{Experimental Setup}
\noindent
\textbf{Datasets.}
The Natural Scenes Dataset (NSD)~\cite{allen2022massive} encompasses fMRI obtained from eight participants, who were exposed to a total of 73,000 RGB images.
This dataset has been extensively employed in numerous studies~\cite{lin2022mind,chen2023seeing,takagi2023high,gu2022decoding,scotti2023reconstructing} for the purpose of reconstructing images perceived during fMRI. Following~\cite{scotti2023reconstructing}, our study utilizes data from subjects 1, 2, 5, and 7, who successfully completed all predetermined trials, and these subjects viewed a set of 10,000 natural scene images, each presented three times. The \texttt{training} set for each subject comprises 8,859 image stimuli and 24,980 fMRI trials, while the \texttt{test} set includes 982 image stimuli and 2,770 fMRI trials. All images and their accompanying captions were derived from MS-COCO~\cite{lin2014microsoft}. It is different from previous methods which employed all subject data to train an omnifit model, our \textsc{BrainGuard} maintains data privacy and simultaneously enables each network to leverage insights from others.
\subsection{Comparison}
\textbf{Evaluation Metrics.} For qualitative evaluation, we visually compare our reconstructed images with the ground-truth images and the results of state-of-the-art methods, as shown in Fig.~\ref{fig:all_sub}. For quantitative evaluation, we present eight different image quality metrics for both low-level and high-level evaluations, following~\cite{scotti2023reconstructing}. Low-level image features provide basic information about the visual content and structure of the image, while high-level features capture semantic information, object relationships, and contextual understanding.
 For low-level evaluation, \texttt{PixCorr} represents the pixel-level correlation, and \texttt{SSIM}~\cite{wang2004image} is the structural similarity. \texttt{Alex}\,(2) and \texttt{Alex}\,(5) denote comparisons of the second and fifth layers of AlexNet~\cite{krizhevsky2012imagenet}, respectively.
 For high-level evaluation, \texttt{Incept}. refers to a comparison of the last pooling layer of Inception-v3~\cite{szegedy2016rethinking}, and \texttt{CLIP} is a comparison of the output layer of the CLIP-Vision model \cite{radford2021learning}. \texttt{EffNet-B} and \texttt{SwAV} are distance metrics is from the EfficientNet-B1 \cite{tan2019efficientnet} and SwAV-ResNet50 \cite{caron2020unsupervised} models, respectively.

\begin{table*}[!bth]
\centering
\resizebox{0.9\textwidth}{!}{
\begin{tabular}{ccc|cccc|cccc}
\toprule
&&&
\multicolumn{4}{c|}{\texttt{Low-Level}} &
\multicolumn{4}{c}{\texttt{High-Level}}\\
\cmidrule{4-11}
\multirow{-2}{*}{}
\multirow{-2}{*}{\textit{Found.}}&
\multirow{-2}{*}{\textit{Inter.}}&
\multirow{-2}{*}{\textit{Advan.}}
&{\texttt{PixCorr}}\!~$\uparrow$ &\texttt{SSIM}\!~$\uparrow$ &\texttt{Alex}\,(2)\!~$\uparrow$ &\texttt{Alex}\,(5)\!~$\uparrow$
&{\texttt{Incept}.}\!~$\uparrow$ &\texttt{CLIP}\!~$\uparrow$ &\texttt{EffNet-B}\!~$\downarrow$ &\texttt{SwAV}\!~$\downarrow$  \\
\midrule
&& & $.163$ & $.238$ &  $88.7\%$ & $90.9\%$  & $83.2\%$ & $86.1\%$ & $.856$ & $.471$ \\
$\surd$& $\surd$&  & $.237$ & $.287$ & $90.3\%$ & $93.8\%$  & $90.5\%$ & $89.3\%$ & $.794$ & $.423$ \\
$\surd$ & & $\surd$ & $.261$ & $.298$ &  $93.3\%$ & $95.9\%$  & $93.4\%$ & $92.5\%$ & $.685$ & $.387$ \\
$\surd$ & $\surd$ & $\surd$ &  $\textbf{.313}$  & $\textbf{.330}$ & $\textbf{94.7}\%$ & $\textbf{97.8}\%$ & $\textbf{96.1}\%$ & $\textbf{96.4}$\% & $\textbf{.624}$ & $\textbf{.353}$ \\
\bottomrule
\end{tabular}}
\caption{\textbf{Synchronization Strategies Ablation}: foundational (\textit{Found.}), intermediate (\textit{Inter.}), and advanced layers (\textit{Advan.}). The results of the first row are from \textsc{BrainGuard} without the hybrid synchronization strategy.}\label{table:ablation}
\vspace{-10pt}
\end{table*}

\textbf{Quantitative Results.}
We compare \textsc{BrainGuard} with seven state-of-the-art methods, namely Mind-Reader~\cite{lin2022mind}, Mind-Vis~\cite{chen2023seeing}, Takagi~\textit{et al.}~\cite{takagi2023high}, Gu~\textit{et al.}~\cite{gu2022decoding}, Brain-Diffuser~\cite{ozcelik2023brain}, MindEye~\cite{scotti2023reconstructing},MindBridge~\cite{wang2024mindbridge} and Psychometry~\cite{quan2024psychometry}. \textsc{BrainGuard} achieves 31.3\%, 33.0\%, 94.7\%, and 97.8\% in the low-level metrics of \texttt{PixCorr}, \texttt{SSIM}, \texttt{Alex}\,(2), and \texttt{Alex}\,(5), respectively. For high-level metrics, \textsc{BrainGuard} secures 96.1\%, 96.4\%, 62.4\%, and 35.3\% in \texttt{Incept}., \texttt{CLIP}, \texttt{EffNet-B}, and \texttt{SwAV}, respectively. It is important to note that the results presented in Table~\ref{table:all_result} are the averages from four subject-specific models, each trained with data from the respective subject. \textsc{BrainGuard} significantly outperforms previous SOTA methods across both low-level and high-level evaluation metrics. For low-level metrics (\texttt{PixCorr}, \texttt{SSIM}, \texttt{Alex}\,(2), and \texttt{Alex}\,(5)), it achieves improvements over Mind-Vis by 23.3
\%, 11.0\%, 22.6\%, and 14.6\%, respectively; and over MindBridge by 16.5\%, 7.1\%, 7.8\%, and 2.5\%. Similarly for high-level metrics (\texttt{Incept}, \texttt{CLIP}, \texttt{EffNet-B}, and \texttt{SwAV}), \textsc{BrainGuard} surpasses Mind-Vis by 17.3\%, 20.2\%, 23\%, and 13.8\%, Brain-diffuser by 8.9\%, 4.9\%, 15.1\%, and 7\%, and MindEye by 4.1\%, 2.1\%, 8.9\%, and 6\%. These advancements are particularly notable given that our method requires training in a single session.

\begin{figure}[!t]
    \centering
\includegraphics[width=\linewidth]{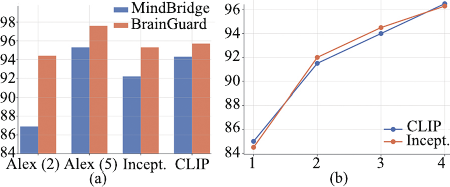}
    \vspace{-5pt}
\caption
{
(a) We evaluate our trained global model and compare it with the SOTA (MindBridge). (b) Performance changes with different numbers of participating subjects.
}\label{fig:combined_figures}
\vspace{-10pt}
\end{figure}

\textbf{Qualitative Results.} As illustrated in Fig.~\ref{fig:all_sub}, the qualitative outcomes align with the quantitative data, indicating that our methodology yields reconstructions of higher quality and greater realism in comparison to alternative approaches. We compare the reconstructions image with MindEye, Psychometry, MindBridge, and Brain-Diffuser, our \textsc{BrainGuard} maintains a
high level of consistency with the visual stimuli in terms of semantics, appearance, and structure. This indicates that our method effectively captures intersubject commonality and individual specificity across subjects, resulting in high-quality image reconstructions from fMRI, and preserving the privacy of individual fMRI.
\subsection{Diagnostic Experiment}\label{sec:ablations}
To systematically demonstrate the influence of each component in \textsc{BrainGuard} on its performance, a series of ablation studies were conducted using the NSD \texttt{test} set.

\textbf{Effective of Hybrid Synchronization Strategy.}
We examined the effectiveness of the hybrid synchronization strategy across foundational, intermediate, and upper layers between the global model and individual models. Table~\ref{table:ablation} displays the experimental results when applying global and individual model synchronization strategies at different levels. For low-level features (such as \texttt{PixCorr} and \texttt{SSIM}), a significant improvement in model performance is observed as we transition to higher layers, particularly when the hybrid synchronization strategy is utilized. For instance, in low-level features, \texttt{Alex}\,(5) achieves a performance of 97.8\%, whereas in high-level features, Inception and CLIP attain performance levels of 96.1\% and 96.4\%, respectively. Additionally, models like \texttt{EffNet-B} and \texttt{SwAV} exhibit varying degrees of improvement when the hybrid synchronization strategy is applied. These findings suggest that the hybrid synchronization strategy effectively enhances the performance of individual models, especially in the intermediate and upper layers.

\textbf{Number of Subjects Participating in Training.} The augmentation in the number of subjects participating in the model training leads to enhanced performance, as each subject can mutually benefit from the shared insights. To empirically verify this premise, we conducted experiments by incrementally increasing the number of subjects involved in the learning process. As shown in Fig.~\ref{fig:combined_figures}(b), our findings highlight a positive correlation between the number of participating subjects and the improvement in results. In particular, performance peaks when all four subjects are trained collaboratively, substantiating the efficacy of our approach. This phenomenon elucidates that with a greater number of subjects engaged in the training process, while each maintains its distinctiveness, the models are adept at identifying and leveraging the shared patterns across different subjects' brain activities, thereby increasing their individual performance.

\textbf{Intersubject Commonality.} As discussed earlier, \textsc{BrainGuard} utilizes a unified training approach on multisubject fMRI, which not only streamlines training costs but also leverages intersubject commonalities. This dual benefit is substantiated through both quantitative and qualitative analyses. Fig.~\ref{fig:combined_figures}(a) presents results comparing our \textsc{BrainGuard}, where the foundational layer in the individual model is followed by the global model’s intermediate and advanced layers, with those obtained using MindBridge. These results demonstrate that \textsc{BrainGuard} has a superior ability to capture intersubject commonalities compared to MindBridge.
Additionally, Fig.~\ref{fig:all_sub} reveals the semantic coherence of reconstructions across various subjects when exposed to the same visual stimuli. This consistency highlights \textsc{BrainGuard}'s proficiency in capturing shared patterns among different subjects, further establishing its effectiveness in fMRI-based image reconstruction.
\section{Conclusion and Discussion}
We introduce \textsc{BrainGuard}, an innovative framework designed to address the challenges of f2I reconstruction, particularly focusing on individual variability and data privacy. By ingeniously integrating individual models for each subject with a global model that captures common and unique patterns of brain activity, \textsc{BrainGuard} ensures a collaborative approach to exploit the intersubject commonalities from the fMRI of multiple subjects while preserving their privacy. A pivotal feature of this framework is the hybrid synchronization strategy, which effectively incorporates insights from the commonality model into personalized models, providing a robust approach for refining image reconstructions.
The deployment of \textsc{BrainGuard} represents a significant advance in brain-computer interfaces, offering a novel and secure method to derive meaningful interpretations from brain.
\section*{Acknowledgments}
This work was supported by the Key R\&D Program of China under Grant No.~2021ZD0112801, the National Natural Science Foundation of China under Grant Nos.~62176108 and U2336212, the Natural Science Foundation of Qinghai Province of China under No.~2022-ZJ-929, the Science Foundation of National Archives Administration of China under No.~2024-B-006, and the Supercomputing Center of Lanzhou University.
\bibliography{main}
\appendix
This supplementary material provides additional information for the AAAI 2025 submission titled ``\textsc{BrainGuard}: \textit{Privacy-Preserving Multisubject Image Reconstructions from Brain Activities.''} The appendix includes implementation details, information on the NSD dataset used, fMRI preprocessing techniques, more details of in DFL module, as well as further reconstruction results, quantitative analyses, broader impacts, and limitations. These topics are organized as follows:
\begin{itemize}
\item[] \S\ref{sec:datasets}: Dataset and fMRI Data Preprocessing Details
\item[] \S\ref{sec:implementation}: Implementation Details
\item[] \S\ref{sec:quantitative}: More Quantitative Results
\item[] \S\ref{sec:more_dfl}: Further Details on the DFL module
\item[] \S\ref{algorithm}: Algorithm of \textsc{BrainGuard}
\item[] \S\ref{sec:more_reconstruction}: More Reconstruction Results
\item[] \S\ref{sec:broaderimpact}: Broader  Social Impacts
\item[] \S\ref{sec:limitation}: Limitations	
\end{itemize}
\section{Dataset and fMRI Preprocessing Details}\label{sec:datasets}
The Natural Scenes Dataset (NSD)\footnote{\url{http://naturalscenesdataset.org/}}\cite{allen2022massive} represents a significant advancement in the field of neuroscience. This comprehensive dataset provides detailed functional Magnetic Resonance Imaging (fMRI) recordings from eight participants. These participants, comprising two males and six females within the age range of 19 to 32 years, were engaged in a unique visual experiment. They viewed a vast collection of 73,000 RGB images, each participant being exposed to a subset of 10,000 unique images, and each image was presented three times to the participants over a series of 20 to 40 sessions.  These sessions comprised whole-brain gradient-echo EPI scans employing 1.8-mm cubic voxels and a repetition time (TR) of 1.6 seconds. Every session was divided into 12 runs lasting 5 minutes each, during which images were presented for 3 seconds followed by a 1-second pause before the next image appeared. Out of the eight participants. Four participants, specifically subjects 1, 2, 5, and 7, completed all sessions.  The images in the NSD were derived from the MS-COCO database~\cite{lin2014microsoft},  formatted into square crops, and presented at a visual angle of $8.4^\circ \times 8.4^\circ$.  While 982 images were shared across all participants, the remaining images were unique to each subject to prevent overlap between their individual sets. Current studies on fMRI-to-image reconstruction~\cite{scotti2023reconstructing,gu2022decoding,takagi2023high} utilizing the NSD dataset generally adhere to a consistent methodology. This approach involves training individual models specific to each of the four participants who completed all scanning sessions. Additionally, these studies utilize a \texttt{test} set comprised of the 982 images that were uniformly presented to every participant. In our Experiments, \textsc{BrainGuard} only training a single session using multi-subject fMRI data while safeguarding the privacy of individual data.

The pre-processing of fMRI data includes conducting temporal interpolation to rectify slice time discrepancies and spatial interpolation to mitigate head motion artifacts. Following this, a general linear model is utilized to calculate single-trial beta weights. Additionally, the NSD dataset incorporates cortical surface reconstructions, which were produced using FreeSurfer\footnote{\url{http://surfer.nmr.mgh.harvard.edu/}}. Both volumetric and surface-based representations of the beta weights are generated, facilitating comprehensive analysis and interpretation. Initially, we masked to the preprocessed fMRI signals using the NSDGeneral Region-of-Interest (ROI) mask, which is specified at a 1.8 mm resolution. The ROI mask encompasses 15724, 14278, 13039, and 12682 voxels for the four subjects respectively, covering various visual areas from the early visual cortex to higher visual areas, and this variation highlights the individual differences in brain structure and function, necessitating the standardization of voxels count across all subjects. 
\section{Implementation Details}\label{sec:implementation}
During the training phase, \textsc{BrainGuard} performs parameter updates using the AdamW optimizer, configured with $\beta_1 = 0.9$, $\beta_2 = 0.9999$, $\epsilon = 10^{-8}$, and a learning rate of $3 \times 10^{-4}$. The \textsc{BrainGuard} model is trained for a total of 600 epochs.
Additionally, during the inference phase, \textsc{BrainGuard} incorporates a retrieval-enhanced inference mechanism. This mechanism involves pre-storing image and text CLIP embeddings as memories during the training phase. The predicted fMRI embeddings, ${\tilde{\bm{I}}}_n$ and ${\tilde{\bm{T}}}_n$, are then used as queries to retrieve the most closely matching CLIP embeddings from these memories based on similarity measures. A weighted average of the retrieved and predicted embeddings is then computed for image reconstruction.
For the reconstruction phase, we follow recent methods~\cite{ozcelik2023brain,quan2024psychometry,wang2024mindbridge} and employ the frozen Versatile Diffusion model~\cite{xu2023versatile}, a multimodal latent diffusion model guided by image and text CLIP embeddings.
\section{More Quantitative Results}\label{sec:quantitative}
\subsection{Number of layer within DFL module.}
We evaluate the layer count $m$ within the DFL module, a hyperparameter in \textsc{BrainGuard}, prompting us to explore its effects through experimentation. As we increment the number of aggregated layers, there is a corresponding increase in the model's computational demands. Through our experimental analysis, where we systematically increased the layer count, we observed a trend in performance variations, which are detailed in Table~\ref{table:layers_num}. In particular, increasing the number of aggregate layers initially leads to improved results; however, a configuration with eight layers emerges as the optimal setting, achieving the highest performance metrics. This finding underscores that the lower layers of the global model predominantly harbor generic information, which is beneficial and sought after by individual models. This insight not only informs the optimal architecture of our dynamic fusion learner module but also highlights the balance between computational efficiency and model performance, guiding future optimizations in \textsc{BrainGuard}'s development.

\begin{table*}[!bth]
\centering
\resizebox{0.9\textwidth}{!}{
\begin{tabular}{c|cccc|cccc}
\toprule
& 
\multicolumn{4}{c|}{\texttt{Low-Level}} & 
\multicolumn{4}{c}{\texttt{High-Level}}\\
\cmidrule{2-9}
\multirow{-2}{*}{Layers}
&{\texttt{PixCorr}}\!~$\uparrow$ &\texttt{SSIM}\!~$\uparrow$ &\texttt{Alex}\,(2)\!~$\uparrow$ &\texttt{Alex}\,(5)\!~$\uparrow$  
&{\texttt{Incept}.}\!~$\uparrow$ &\texttt{CLIP}\!~$\uparrow$ &\texttt{EffNet-B}\!~$\downarrow$ &\texttt{SwAV}\!~$\downarrow$  \\
\midrule
 $12$ &  $.245$  &  $.306$  &   $90.5$\%  &  $93.8$\%   &  $83.2$\%  &  $86.2$\%  &  $.826$  &  $.459$  \\
 $10$ &  $.257$  &  $.324$  &   $93.5$\%  &  $95.3$\%   &  $93.9$\%  &  $92.9$\%  &  $.675$  &  $.374$  \\
 $8$  & $\textbf{.313}$  & $\textbf{.330}$ & $\textbf{95.2}\%$ & $\textbf{98.1}\%$ & $\textbf{96.3}\%$ & $\textbf{96.5}$\% & $\textbf{.624}$ & $\textbf{.354}$ \\
 $6$  &  $.275$  &  $.319$  &   $93.9$\%  &  $96.2$\%   &  $93.1$\%  &  $93.0$\%  &  $.664$  &  $.363$ \\
 $4$  &  $.264$  &  $.321$  &   $94.2$\%  &  $96.3$\%   &  $93.5$\%  &  $93.2$\%  &  $.657$  &  $.359$ \\
 $2$  &  $.257$  &  $.324$  &   $93.5$\%  &  $95.3$\%   &  $91.4$\%  &  $92.9$\%  &  $.675$  &  $.374$ \\
 $1$  &  $.245$  &  $.335$  &   $90.5$\%  &  $93.8$\%   &  $89.2$\%  &  $89.6$\%  &  $.726$  &  $.439$\\
\bottomrule
\end{tabular}}
\caption{Ablation study focusing on the impact of varying the number of trainable parameters within the DFL module.}\label{table:layers_num}
\vspace{-10pt}
\end{table*}
\section{Further Details on the DFL module}\label{sec:more_dfl}
We update the learnable weights $W_{s}^{m}$ of the advanced $m$-layer parameters using the following Eq.~\eqref{update_weight_more}:
\begin{equation}\label{update_weight_more}
\begin{split}
W_{s}^{m} &\leftarrow W_{s}^{m} - \eta \frac{\partial \mathcal{L}(\theta_s; \theta_{\rm g})}{\partial W_{s}^{m}}\\
&= W_{s}^{m} - \eta \frac{\partial \mathcal{L}(\theta_s; \theta_{\rm g})}{\partial \theta_{s}^{m}} \odot \frac{\partial \theta_{s}^{m}}{\partial W_{s}^{m}} \\
&= W_{s}^{m} - \eta \frac{\partial \mathcal{L}(\theta_s; \theta_{\rm g})}{\partial \theta_{s}^{m}} \odot (\theta_{\rm g} - \theta_s)^m,
\end{split}
\end{equation}
where $\theta_{s}^{m}$ denotes the advanced $m$-layer parameters of the individual model $\theta_s$, and $(\theta_{\rm g} - \theta_s)^m$ represents the difference in the advanced $m$-layer parameters between the global model $\theta_{\rm g}$ and the individual model $\theta_s$. Note that during this phase, both $\theta_{\rm g}$ and $\theta_s$ are kept frozen.

\begin{figure}[!t]
\centering
\includegraphics[width=0.95\linewidth]{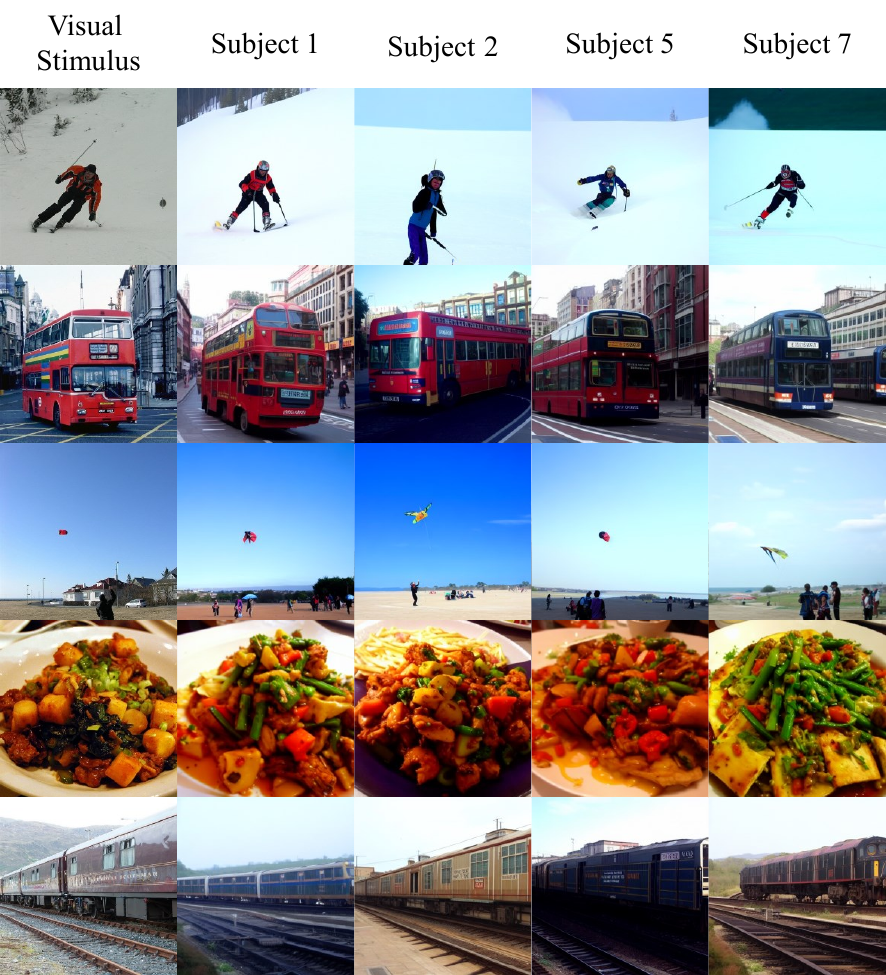}
\caption{Examples of fMRI reconstructions generated by our \textsc{BrainGuard} model are presented. The first column shows the visual stimulus (ground truth) image in the NSD \texttt{test}, while the remaining columns correspond to individual subjects (Sub1, Sub2, Sub5, Sub7).}
\label{fig:all_sub}
\end{figure}

\section{Algorithm of \textsc{BrainGuard}}\label{algorithm}
The procedure of \textsc{BrainGuard} is outlined in Algorithm~\ref{algo1}.
\begin{algorithm}[!t]
\label{algo1}
\renewcommand{\algorithmicrequire}{\textbf{Input:}}
\renewcommand{\algorithmicensure}{\textbf{Output:}}
\caption{\textsc{BrainGuard} algorithm.}
\label{algo1}	
\begin{algorithmic}[1]
\REQUIRE $\{\bm{X}_{s,n}|\,\forall\,s\in\{1,\ldots,S\}, \forall\,n\in\{1,\ldots,N\}\}$\,. 
\ENSURE Optimized parameters: $(\theta_1^\star,\ldots,\theta_S^\star)$\,.
\STATE \textbf{Initialization:}  $W_s^m, \theta^0_s,\theta^{'0}_s, \theta^0_{\rm g}$, $\alpha$ and $epoch_{\max}$\,.
\FOR{$t\in\{1,\ldots,epoch_{\max}\}$}
\FOR{$s\in\{1,\ldots,S\}$}
\STATE Obtain $W_s^m$ by training Eq.~\eqref{update_weight_more}\,;
\STATE $\theta_s^t\leftarrow\theta_s^{t-1} + (\theta_{\rm g}^{t-1} - \theta_s^{t-1}) \odot[\bm{1}^{\left| \theta_s \right|-m}; W_s^m]$\,;
\FOR{mini-batch samples in $\{\bm{X}_{s,n}\}$}
\STATE Update $\theta_s$ by minimizing $\mathcal L_s$\,;
\STATE $\theta'_s \leftarrow \alpha \theta'_s + (1-\alpha) \theta_s$\,;
\ENDFOR
\ENDFOR
\STATE Update $\theta^t_{\rm g}$ by $\theta_{\rm g}^t = \sum_s k_s \theta_s^{'t}$, where $k_s = \left| \bm{X}_{s,n} \right| / \sum^{S}_{i=1}\left| \bm{X}_{i,n} \right|$\,;
\ENDFOR
\end{algorithmic}
\end{algorithm}
\section{More Reconstruction Results}\label{sec:more_reconstruction}
Figure~\ref{fig:all_sub} demonstrates the semantic coherence and visual variations in the reconstruction results across different subjects when presented with the same visual stimuli. This consistency underscores \textsc{BrainGuard}'s proficiency in identifying common patterns among subjects while preserving individual-specific traits and maintaining the confidentiality of brain data during fMRI-based image reconstruction. The observed variations further emphasize the inherent subject-specific nature of fMRI data.

\begin{figure*}[!t]
\centering    
\includegraphics[width=0.90\linewidth]{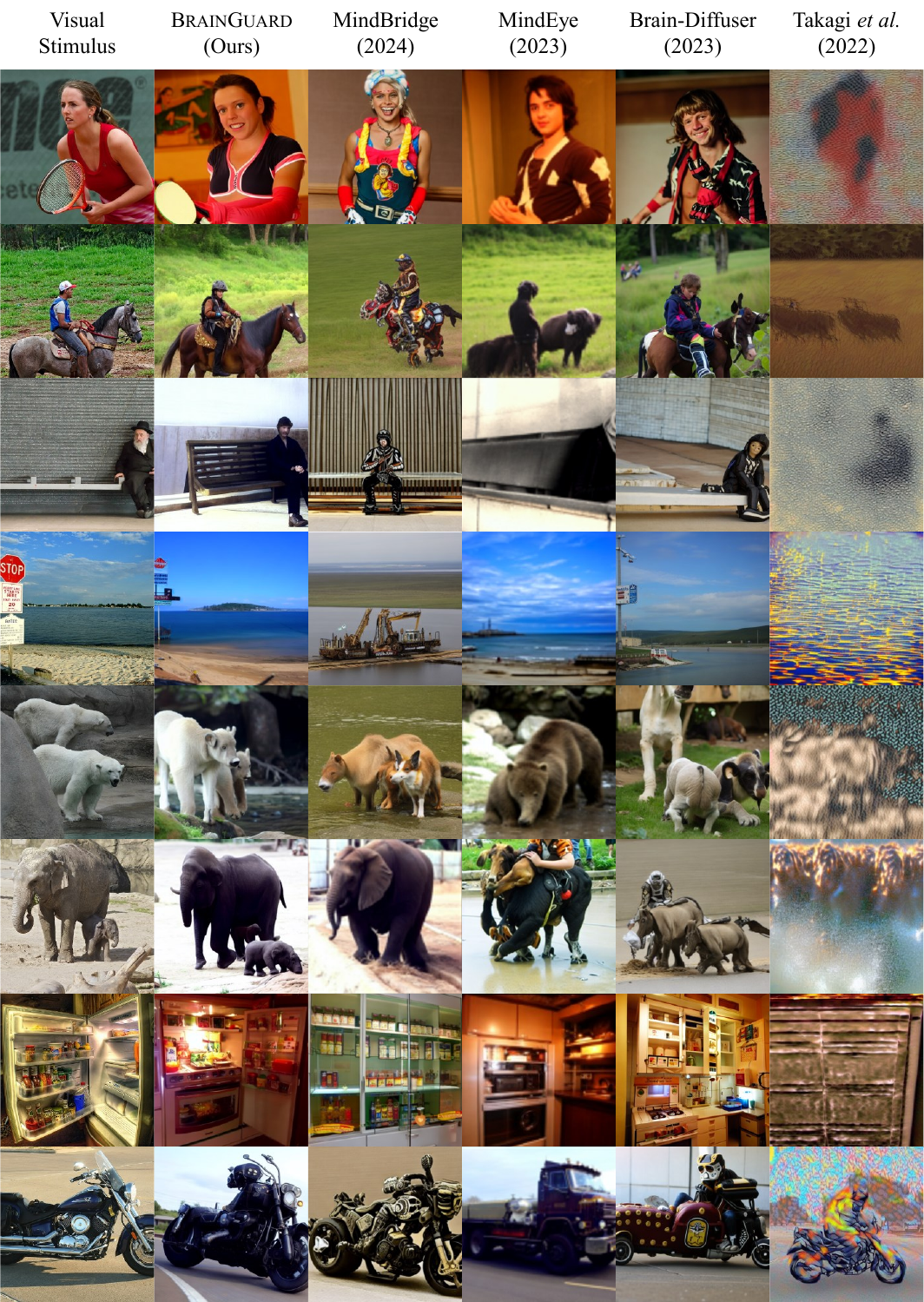}
\caption{Additional resconstruction results on NSD \texttt{test} are compared with other state-of-the-art (SOTA) methods.}
\label{fig:vis_supp}
\end{figure*}

Additionally, the visual results in Fig.~\ref{fig:vis_supp} showcase \textsc{BrainGuard}'s ability to generate reconstructions of high quality and realism. Although these reconstructions are not exact duplicates of the original stimulus images, they effectively retain most of the layout and semantic content.
\section{Broader Social Impacts}\label{sec:broaderimpact}
This paper introduces \textsc{BrainGuard}, a privacy-preserving collaborative framework for reconstructing images from human brain activity via multisubject fMRI data. This framework is a major leap forward in the Brain-Computer Interface domain.
\textsc{BrainGuard} represents a significant advancement, holding the promise to vastly enhance our comprehension of brain functionality. \textsc{BrainGuard} is distinguished by its ability to adeptly capture the communal and individualized aspects of brain activity, thereby contributing to the personalization of medical approaches. Moreover, it emphasizes the protection of privacy and security concerning the sensitive brain data it processes. Looking ahead, technologies capable of decoding human brain activity are expected to revolutionize our interaction with digital environments, introducing novel interfaces that facilitate direct communication with human cognition.
\section{Limitations}\label{sec:limitation}
Currently, \textsc{BrainGuard} is specifically designed for the purpose of image reconstruction using fMRI data. This opens a significant opportunity to refine our methodology to include more complex human brain activity signals, such as those derived from magnetoencephalography (MEG) and electroencephalography (EEG). Furthermore, \textsc{BrainGuard}, consistent with prior research, primarily focuses on decoding signals from the visual cortex. Nonetheless, considering the complexity of human vision as a cognitive process, which involves contributions from beyond the visual cortex, it is crucial for forthcoming research to extend its analysis to additional brain regions. Such an expansion would not only necessitate the development of more sophisticated algorithms capable of parsing and interpreting a more complex array of neural signals but also require a conceptual shift towards viewing vision as a multi-faceted cognitive process that integrates sensory input with memory, emotion, and decision-making processes.
\end{document}